\documentclass[sigconf, 10pt]{acmart}

\usepackage[T1]{fontenc}

\usepackage{latexsym}
\usepackage{tikz}
\usepackage{tkz-graph}
\usepackage{subcaption}
\usetikzlibrary{arrows,automata,arrows.meta}

\usepackage{microtype}

\newcommand\BibTeX{B\textsc{ib}\TeX}

\usepackage{multirow}
\usepackage{ctable}
\usepackage{xcolor}

\usepackage{url}
\usepackage{capt-of}

\usepackage[utf8]{inputenc}
\usepackage[T2A, T1]{fontenc}

\usepackage{cleveref}

\crefformat{section}{\S#2#1#3}
\crefformat{subsection}{\S#2#1#3}
\crefformat{subsubsection}{\S#2#1#3}
\crefrangeformat{section}{\S\S#3#1#4 to~#5#2#6}
\crefmultiformat{section}{\S\S#2#1#3}{ and~#2#1#3}{, #2#1#3}{ and~#2#1#3}

\usepackage{graphicx}
\graphicspath{ {./images/} }

\AtBeginDocument{%
  \providecommand\BibTeX{{%
    \normalfont B\kern-0.5em{\scshape i\kern-0.25em b}\kern-0.8em\TeX}}}

\acmConference[Workshop on Multilingual Search]{Workshop on Multilingual Search, The Web Conference}{WWW}{April 2021}

\begin{document}

\title[Instance-based Transfer Learning for Multilingual Deep Retrieval]{Instance-based Transfer Learning for \\Multilingual Deep Retrieval}

\author{Andrew O. Arnold}
\authornote{Work done while at Google Research.}
\affiliation{%
  \institution{AWS AI}}
\email{anarnld@amazon.com}

\author{William W. Cohen}
\affiliation{%
  \institution{Google Research}}
\email{wcohen@google.com}

\renewcommand{\shortauthors}{Arnold and Cohen}

\begin{abstract}

We focus on the problem of search in the multilingual setting.  Examining the problems of \emph{next-sentence prediction} and \emph{inverse cloze}, we show that at large scale, instance-based transfer learning is surprisingly effective in the multilingual setting, leading to positive transfer on all of the 35 target languages and two tasks tested.  We analyze this improvement and argue that the most natural explanation, namely direct vocabulary overlap between languages, only partially explains the performance gains: in fact, we demonstrate target-language improvement can occur after adding data from an auxiliary language even with no vocabulary in common with the target. This surprising result is due to the effect of transitive vocabulary overlaps between pairs of auxiliary and target languages.

\end{abstract}

\keywords{multilingual, neural networks, transfer learning, information retrieval}

\maketitle

\section{Introduction}

In this paper we explore the problem of search in the multilingual setting.  Specifically, we analyze the behavior of instance-based transfer
learning on two very large-scale deep retrieval tasks (\emph{next-sentence prediction} and \emph{inverse cloze}) across dozens of auxiliary languages. Motivated by the increasing availability of unlabelled data of various sizes across languages, we introduce a simple technique for combining vocabularies across languages.  We show that this method is surprisingly effective, leading to positive transfer on all of the 35 target languages tested across both tasks, and relative improvements of up to 200\%.  

Analysis of this result reveals a number of influences on transfer-learning performance.  Unsurprisingly, performance is improved more when the task is difficult, when the available target language data is limited and when there is a large overlap between target and auxiliary language vocabularies.   However, analysis suggests that these effects are only a partial explanation of the effectiveness of instance-based transfer. To support this argument, we demonstrate that multilingual instance-based transfer can lead to target-language improvement after adding data from an auxiliary language with no vocabulary in common with the target: this surprising result is due to the effect of transitive vocabulary overlaps between pairs of auxiliary and target languages.

We conclude by contextualizing these results in the broader literature of multilingual search and transfer learning, and arguing that the results are important due to their generalizability across models and architectures, limited only to the wide class of settings where raw, unaligned text across languages is available.

\section{Background and Related Work}
    \label{sec:background}
    
    In \emph{deep retrieval} (or \emph{deep document linking}), the goal is to learn to retrieve relevant documents
from a large corpus of candidates, based on similarity to a query
document.  In \emph{multilingual transfer learning}, we wish to improve
performance on some task by using data from auxiliary languages to
improve performance on a designated target language.  Perhaps the
simplest type of multilingual transfer learning is \emph{instance-based
transfer learning}, in which, for example, data from the target language and the
auxiliary languages may be pooled, and a single model learned from the
pooled data \cite{lin2013double}.  It is not immediately obvious when instance-based
transfer learning will improve performance in this multilingual
setting: for instance, a plausible conjecture is that pooling data in
this way would only improve performance if the amount of auxiliary data
was carefully balanced with the amount of target data, or if the
auxiliary languages were carefully selected to be highly similar to the
target.

\subsection{Next-Sentence Prediction and Inverse Cloze}
    \label{sec:model}
    
    We examine deep retrieval on the tasks of next sentence prediction (NSP) and inverse cloze (IC).  In NSP the goal is to identify the next sentence in a document from among a corpus of hundreds of millions of candidate sentences across dozens of languages, given only the current sentence as context.  NSP has many important applications to areas such as question answering \cite{rajpurkar-etal-2018-know}, language model training \cite{devlin2018bert}, summarization \cite{liu2019comes} and conversational modeling \cite{vinyals2015neural}.  IC is a slight generalization of NSP, where instead of predicting the single next sentence given a query sentence, the model must now return the entire context surrounding that sentence.  Considering the query sentence as a \emph{question} and its context as a passage containing a potential \emph{answer}, this task has a direct relationship to open domain question answering \cite{DBLP:journals/corr/abs-1906-00300, doi:10.1177/107769905303000401}.

    While NSP and IC have been well studied from the perspective of sequence modeling \cite{ghosh2016contextual} and binary classification \cite{devlin2018bert}, we follow an alternative line of work that models the problems as instances of deep retrieval in extremely large output spaces \cite{logeswaran2018an, 46057, pmlr-v89-reddi19a}.  Specifically, we generate a shared set of unigram and bigram features representing the \emph{current} and \emph{next} (\emph{surrounding}) sentences in an NSP (IC) pair, respectively.  We then train a feed-forward neural network that learns to maximize the dot-product between these consecutive sentences (contexts), as represented by the learned embedding vectors of each of the shared vocabulary's n-gram tokens.  We follow the architecture\footnote{Further reproducibility details such as specific hyperparameter, optimization and inference settings are unfortunately unavailable due to change of employers.} of \citet{10.5555/2987189.2987282, logeswaran2018an, 46057, pmlr-v89-reddi19a}, using a siamese network to learn the multilingual vocabulary embeddings end-to-end, along with our NSP and IC objective functions, with all embeddings across languages mapped into a single shared space.

\subsection{Multilingual Deep Retrieval}
    \label{sec:multiling}

    This architecture allows us to efficiently use an extremely large vocabulary of simple unigrams and bigrams that more computationally intensive techniques do not currently permit.  Using such large vocabularies, these models are able to identify and exploit tiny cross-lingual dependencies found among the many tail n-grams observed in the large unsupervised monolingual datasets available in many languages across the internet.  This follows other work that leverages the concatenation of monolingual corpora in the multilingual setting \cite{devlin2018bert, johnson-etal-2017-googles}, and is in contrast to other work that tries to learn multilingual embeddings via attention \cite{logeswaran2018an, DBLP:journals/corr/abs-2005-00633, feijo2020mono, pires2019multilingual, wu2019beto, wu-dredze-2020-languages}, shared cross-lingual embedding spaces \cite{chen-cardie-2018-unsupervised, arivazhagan2019massively, mikel2020on, karthikeyan2019cross, Cao2020Multilingual}, cross-lingual mapping functions \cite{xu-etal-2018-unsupervised-cross, kulshreshtha2020crosslingual}, other outside structure \cite{wang-etal-2013-transfer,10.1007/978-3-642-25073-6_42, mueller-etal-2020-analysis} and domain knowledge \cite{plank-agic-2018-distant, koehn-knight-2002-learning}.

    For language modelling and question answering, the NSP and IC objectives are interesting because they lead to difficult classification tasks for which labeled data is plentiful.  The motivation for using these tasks here is similar: under the deep retrieval formulation, NSP and IC are difficult, because they require comparison against all possible candidate sentences and contexts in the corpus, forcing the model to learn very nuanced distinctions, and potentially generating embeddings that generalize well to downstream tasks.

\subsection{Instance-based Transfer Learning}

    NSP and IC are attractive problems for unsupervised and semi-supervised transfer learning in part because of the large amount of otherwise unlabelled, yet ordered, language data available on the internet.  Even for such unlabelled data, there is still more data available in certain languages than in others, making this problem also an attractive setting for studying transfer learning between high and low resource languages \cite{5288526, Silver1998}.

    We focus on \emph{instance-based transfer learning} where we use examples drawn from a related but distinct \emph{auxiliary} dataset to improve performance on a \emph{target} dataset \cite{Wang2018InstanceBasedDT}.  However, rather than trying to align embeddings learned across languages \cite{sogaard-etal-2018-limitations, nakashole-2018-norma, alvarez-melis-jaakkola-2018-gromov}, we instead attempt the easier task of first aligning vocabulary items across languages (by simply matching identical tokens), and then learning a single shared embedding for each vocabulary item.

\subsection*{Instance-based transfer vs. fine-tuning}

    Instance-based transfer is similar to another widely-used neural transfer method, namely \emph{fine-tuning}.
    In \emph{fine-tuning}-based transfer learning \cite{howard2018universal, 6639081, 6424230, 6288862}, the weights of a network are trained on one (large) set of auxiliary data, and then copied into a new network where they are further adjusted using the target data.  So in fine-tuning the model is optimized twice---once on the auxiliary data, and once on the target data---whereas in instance-based transfer it is optimized once, jointly, on both auxiliary and target datasets.
    
    While fine-tuning allows additional flexibility, since one can independently decide optimization hyperparameters for each pass, this flexibility comes with some costs.  In our setting, with 35 potential target languages (each with 34 potential auxiliary languages), instance-based learning produces a single model; while fine-tuning would produce 35, one specialized for each language, and in practical settings each of these 35 different models would need to be separately stored, maintained, etc.
    The sequential nature of fine-tuning also means that there are many choices to make when there are multiple auxiliary languages: for instance, in learning an NSP model for Ukrainian, perhaps it is better to train first on English (the most frequent language), then Russian (a more frequent related language), and finally fine-tune on Ukrainian.  In order to fully utilize all potential auxiliary languages, one would have to consider fine-tuning the model up to 34 different times!  In this broader setting, fully exploring the space of fine-tuned transfer models is a substantial undertaking.
    
    In contrast, with the instance-based transfer approach we train a single joint model which shows improvements across all languages.  If necessary, this efficiently computed multi-domain instance-based transfer model could be further refined by more expensive target-dependent fine-tuning; however, we leave exploration of such approaches for future work, focusing here on careful study of the more efficient instance-based method, at large scale, across two tasks and many target languages.

\section{Experiments}

    \subsection{Next Sentence Prediction}
    
\begin{figure}[htbp]
            \centering
            \includegraphics[scale=.65]{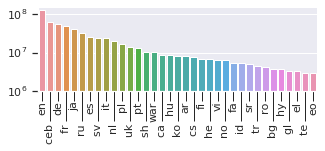}
            \setlength{\belowcaptionskip}{-15pt}
            \caption{Train sentence pairs per language (log scale).}
            \label{fig:lang_cdf}
        \end{figure}

        We extract approximately 720 million next sentence pairs from publicly available Wikipedia\footnote{Downloaded May 11 and December 2, 2019.}, restricting our experiments to the top 35 languages\footnote{We use the same language code abbreviations as the Wikipedia subdomains associated with each language (e.g., \emph{\textbf{en}.wikipedia.com} for English), and the special code \emph{\textbf{all}} for the combined dataset comprised of data from all languages. Chinese is excluded due to the lack of a suitable tokenizer.} which account for 90\% of the data.  We split this dataset into train, development, and evaluation splits of 90\%, 5\% and 5\% respectively.  Figure \ref{fig:lang_cdf} shows the relative size of each language in the dataset.
        
        For each section in each article in each language we extract a pair of consecutive sentences if both sentences have at least four words.  We then use a bag-of-n-grams representation for each sentence, constructing a training example as the unigram and bigram features from each $\langle current, next \rangle$ sentence pair, and aggregate all unique unigrams and bigrams into a shared \emph{vocabulary}.  This results in per-language vocabularies ranging from 4 to 200 million n-gram tokens.  For the special collection \emph{all}, containing the union of all tokens across all languages, we limit the vocabulary size to the top 350 million n-gram tokens across all languages, sorted by frequency.

\begin{figure}[htbp]
            \centering
            \includegraphics[scale=.40]{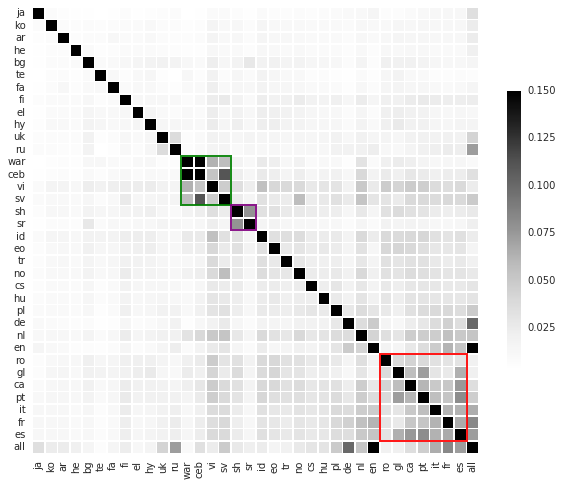} 
            \caption{Jaccard index of vocabularies across languages, with blocks highlighted for Austronesian/Lsjbot (green), Serbian/Serbo-Croatian (purple), and Romance (red) languages.}
            \label{fig:vocab_jac}   
        \end{figure}

        These 35 languages include some pairs that are quite similar and some pairs that are quite different.  To give a rough measure of this, we looked at the similarity of the vocabularies.  Figure \ref{fig:vocab_jac} shows the \emph{Jaccard index} among the vocabularies of the languages within the corpus, a measure of vocabulary similarity defined for a pair of language vocabularies $V_a$ and $V_b$ as $\frac{|V_a \cap V_b|}{|V_a \cup V_b|}$.  The matrix has been sorted to emphasize clusters roughly corresponding to known language groups (e.g., the Romance languages  \emph{ro-gl-ca-pt-it-fr-es}).  Other interesting structure observed includes the large overlap between Serbian and Serbo-Croatian (\emph{sr-sh}); and the cluster of Cebuano and Waray (two Austronesian languages spoken in the Philippines) with Vietnamese and Swedish (\emph{ceb-war-vi-sv})\footnote{Most of the Cebuano and Waray articles were written by a computer program, \emph{Lsjbot}, which has also written articles for Swedish Wikipedia, accounting for the large unexpected overlap in vocabularies.}.       
        
\begin{figure*}
\vspace{-8pt}
\begin{subfigure}{.5\textwidth}
  \centering
  \includegraphics[width=1.0\linewidth]{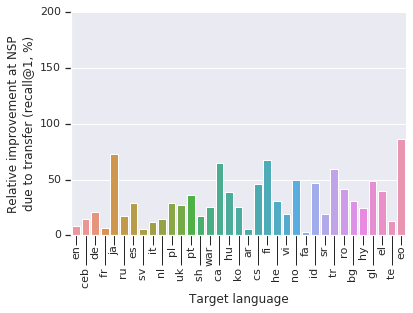}
  \label{fig:lang_xfer_nsp}
\end{subfigure}%
\begin{subfigure}{.5\textwidth}
  \centering
  \includegraphics[width=1.0\linewidth]{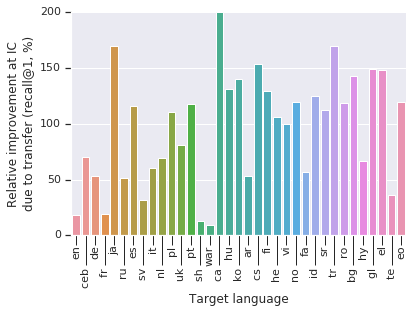}
  \label{fig:lang_xfer_ic}
\end{subfigure}
\vspace{-24pt}
\setlength{\belowcaptionskip}{-10pt}
\caption{Relative transfer improvement on each target language for NSP (left) and IC (right) tasks.  Languages are sorted by number of training sentences, as in Figure \ref{fig:lang_cdf}.}
\label{fig:lang_xfer}
\end{figure*}

\vspace{-18pt}

    \subsection{Inverse Cloze}
    
    We use the same Wikipedia corpus for the inverse cloze task, this time randomly sampling a single sentence from each section in each article in each language as our $current$ sentence, and concatenating the surrounding four sentences (two preceding and two following) as our $context$, resulting in approximately 60 million $\langle current, context \rangle$ pairs.  Feature construction, model architecture, training and evaluation are the same as in NSP.

    \subsection{Method}
    
        Using the architecture described in Section \ref{sec:model}, we train two different types of models: a \textbf{per-language} model trained on a corpus containing only sentence pairs obtained from a particular language's Wikipedia subdomain; and a \textbf{combined} model trained on a corpus containing the aggregated training examples across all languages.  We train one per-language model for each language, along with a single combined model, resulting in 36 models total.  We evaluate each per-language model's performance on that language's held-out evaluation data, and we evaluate the single combined model's performance likewise on each individual language's evaluation data, resulting in 35 pairs of evaluation data.  We evaluate the performance of these models using a sampled \textbf{recall@k} metric defined as the percent of query sentences for which the correct next sentence (context) is retrieved within the top-k results\footnote{To determine the top-k results we combine the true next sentence (context), drawn from the evaluation data, with a sample of 100,000 false next sentence (context) distractors, drawn from the training data.  All 100,001 sentences (contexts) are embedded in the same space and their similarity computed relative to the test query, with the top-k most similar sentences (contexts) retrieved.}.  We define the \textbf{relative transfer improvement} as:
        
      
        \mathchardef\mhyphen="2D
        \[
        \frac{recall@k_{combined} - recall@k_{per\mhyphen language}}{recall@k_{per\mhyphen language}}
        \]

        
\section{Results}

    Figure \ref{fig:lang_xfer} shows the relative transfer improvement\footnote{Baseline performance is shown in the x-axes of Figure 5.} for each language, for each task, on the recall@1 metric.  Perhaps surprisingly, we observe that all languages, across both tasks, benefit from the addition of data from other languages (i.e., there is no observed negative transfer, in contrast with the observations of \citet{arivazhagan2019massively}), with most languages being improved by over 25\%.  This lack of negative transfer could be due to the manner in which transfer is being achieved, with only examples from the target domain that reinforce the source domain being picked-up by the model (see \cref{sec:direct_overlap,sec:indirect_overlap}).
    
    We also see that the average relative benefit of transfer on the IC task (96\%) is over three times greater than on the NSP task (31\%).  Despite this difference in the magnitude of the effect, the Pearson correlation coefficient of the relative effect for each language across tasks is quite high at .77.  This suggests there are per-language effects, persisting across tasks, that at least somewhat influence the effectiveness of transfer.  We investigate potential explanations for these effects below.

         \begin{figure}[h]
            \centering
            \includegraphics[scale=.55]{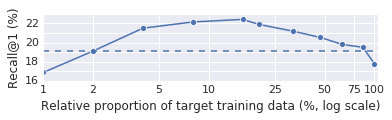}
            \setlength{\belowcaptionskip}{-10pt}
            \caption{Performance on Ukrainian data as a function of the \emph{mixture ratio} between \emph{target} data (Ukrainian) and  \emph{auxiliary} data (\textbf{all} excluding Ukrainian) on NSP.  Dashed line shows baseline performance using \emph{native} Ukrainian mixture ratio of 2\%.}
            \label{fig:xfer_ratio}
        \end{figure}

                    \begin{figure*}[htp]
\begin{subfigure}{.5\textwidth}
  \centering
  \includegraphics[width=1.0\linewidth]{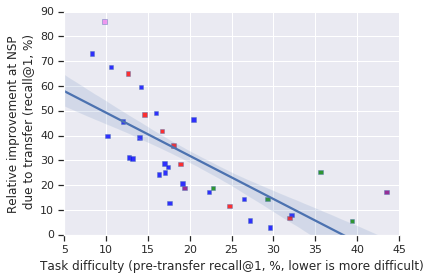}
  \label{fig:xfer_difficulty_nsp}
\end{subfigure}%
\begin{subfigure}{.5\textwidth}
  \centering
  \includegraphics[width=1.0\linewidth]{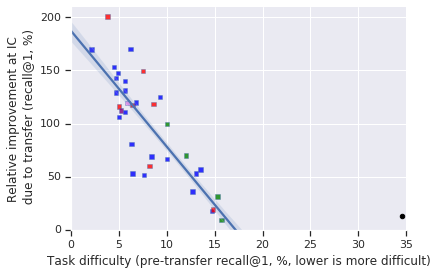}
  \label{fig:xfer_difficulty_ic}
\end{subfigure}
\vspace{-18pt}
\caption{Relative transfer improvement for NSP (left) and IC (right) as a function of task difficulty (as measured by pre-transfer recall@1), along with linear fits $\pm$ one standard deviation ($R^2$ = 53\% and 70\%, respectively).  Green, purple and red points correspond to clusters in Figure \ref{fig:vocab_jac}, violet is Esperanto.  The outlier in the IC plot (Serbo-Croatian (\textbf{sh}), marked in black) has been removed from the linear fit to better show the trend.}
\label{fig:xfer_difficulty}
\end{figure*}   

   \subsection{Performance Improves for Many Mixture Ratios}

    These results are surprisingly strong for such a simple method.
    It is natural to ask if instance-based transfer can be improved, for instance
    by weighting the auxiliary data differently.  In the experiments above, we 
    built a \emph{combined} model with relative language proportions unchanged from their original ratio in the raw Wikipedia data.  
    Reasoning that some amount of transfer from the auxiliary data might be beneficial, but too much transfer might overwhelm the target language, Figure \ref{fig:xfer_ratio} shows the results for a particular language (Ukrainian) of adjusting this \textbf{mixture ratio} between the target language (Ukrainian) and auxiliary data (all languages except Ukrainian) on the NSP task.  
    We observe that performance increases as the amount of target data used increases up to a point, and then starts to diminish, with an optimum mixture ratio of 10 - 20\%.  In all of these experiments, the total amount of data seen by the model is unchanged, only the ratio of the sources varies.  
    In summary, gains are observed for a wide range of mixing ratios, but the chosen ratio is
    not optimal: 
    there is a potential relative improvement of 17\% over the \emph{native} Ukrainian mixture ratio of 2\% found in the original corpus, suggesting further improvement to the overall performance of the combined model could be achieved by optimizing all the mixture ratios across different languages.
    Even though there are clearly additional improvements to be obtained
    by tuning instance-based transfer, a more fundamental question to ask is why the method
    works so well. The following sections explore this question further, by investigating potential explanations for when and why this instance-based transfer improvement occurs.

    \subsection{Sample Size: Low-resource Languages Improve More}  \label{sec:size} 
    
        \begin{figure}[h]
        
            \centering
            
            \includegraphics[scale=.52]{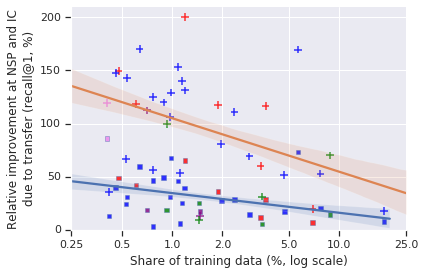}
        
            \setlength{\belowcaptionskip}{-15pt}
            
            \caption{Relative transfer improvement for NSP (\rule{0.5em}{0.5em}, blue fit line) and IC ($+$, orange fit line) as a function of training data sample size, along with linear fit $\pm$ one standard deviation ($R^2$ = 14\% and 18\%, respectively).  Green, purple and red points correspond to clusters in Figure \ref{fig:vocab_jac}, violet is Esperanto.}
            \label{fig:xfer_size}
            
        \end{figure}

        Figure \ref{fig:xfer_size} shows the relative transfer improvement on both tasks for each language on the recall@1 metric, as a function of that language's share of the training data.  Unsurprisingly, we observe that the languages with the smallest share of training data receive the largest improvement from this transfer.  Interestingly, the language with the largest relative NSP improvement is Esperanto, a constructed language with vocabulary drawn from Romance and Germanic languages.

    \subsection{Task Difficulty: Harder Tasks Improve More} 
    
        Figure \ref{fig:xfer_difficulty} shows the relative transfer improvement for each language on both tasks, as a function of each language's baseline, pre-transfer, performance on each task.  In other words, for each language, we measure the performance of a model trained only on that language, and use that as a proxy for the difficulty of the task: the lower this baseline score, the harder the underlying task.  Across both tasks we see a very strong positive relationship between pre-transfer task difficulty (measured as pre-transfer recall@1) and post-transfer relative improvement:  the more difficult the initial task is, the more room there seems to be for improvement from transfer.

        Combined with the results from \S~\ref{sec:size} relating transfer improvement and sample size, we see it is not just the lack of available data or the difficulty of the underlying task that makes transfer effective, but it is actually the combination of the two.  Transfer seems to help when \emph{performance} at a task can be improved by \emph{adding} auxiliary data.  When baseline \emph{performance} is low, there is more room for improvement.  Likewise, when baseline data is small, there is more room for \emph{adding} information from other related sources.  If the performance is already high, or if the target data is already sufficient to saturate the model, there is less opportunity for transfer to make a difference.

    \subsection{Direct Vocabulary Overlap}
        \label{sec:direct_overlap} 
        \begin{figure}[htbp]
            \centering
            \includegraphics[scale=.52]{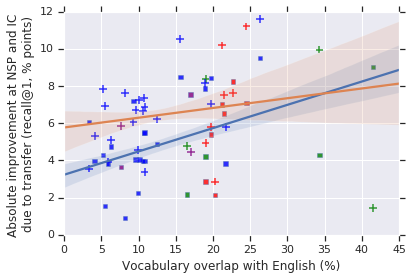}
            \caption{Absolute transfer improvement for NSP (\rule{0.5em}{0.5em}, blue fit line) and IC ($+$, orange fit line) as a function of vocabulary overlap with \textbf{en}, along with linear fit $\pm$ one standard deviation ($R^2$ = 23\% and 4\%, respectively).  Green, purple and red points correspond to clusters in Figure \ref{fig:vocab_jac}.}
            \label{fig:xfer_vocab}
        \end{figure}        
        
    In addition to the observed relationships with inverse sample size and task difficulty, transfer also seems to improve when there is larger overlap between language vocabularies.  Using English as a reference language, Figure \ref{fig:xfer_vocab} shows the absolute transfer improvement in recall@1 for each language on both tasks as a function of that language's vocabulary's overlap with the English vocabulary $\frac{|V_{target} \cap V_{English}|}{|V_{target}|}$.  The larger the overlap, the larger the benefit from transfer.  This makes sense in light of the model architecture, as it is the presence of overlapping tokens from different languages in the same training sentence pairs that allows information learned on one language's data to impact another language.  Combined with the significant joint overlap among language vocabularies shown in Figure \ref{fig:vocab_jac}, this helps demonstrate why transfer can occur even between seemingly unrelated languages.

Table \ref{tab:variance} summarizes the explanatory effects of \textbf{sample size}, \textbf{task difficulty} and \textbf{vocabulary overlap} on the effectiveness of transfer across the NSP and IC tasks.  These factors combined explain 62\% and 78\% (respectively) of the variance of transfer improvements across both tasks ($R^2$).  Even accounting for these observed effects, however, there is still a significant amount of unexplained variance in the observed transfer improvement across languages, suggesting some other effect may be at work.
    
    \begin{table}[htp]
        \begin{tabular}{|l|c|c|c|}
        \hline & \bf NSP & \bf IC & \bf Average\\ \hline
        \textbf{sample size} & 14\% & 18\% & 16\% \\
        \hline
        \textbf{task difficulty} & 53\% & ~70\%\footnotemark & 62\% \\
        \hline
        \textbf{vocabulary overlap} & 23\% & ~4\% & 14\% \\
        \hline
        \hline
        \emph{combined} & 62\% & 78\% & 70\% \\
        \hline        
        \end{tabular}    
        \vspace{6pt}
        \caption{Percent of observed relative transfer improvement variance explained by various factors ($R^2$) across NSP and IC tasks.}
        \vspace{-12pt}
        \label{tab:variance}
        \end{table}    
    
    \footnotetext{Serbo-Croatian removed.}
    
    \subsection{Transitive Vocabulary Overlap}
    \label{sec:indirect_overlap} 

\begin{table*}[ht]
\begin{minipage}[b]{0.47\linewidth}
\centering


        \centering

        \begin{center}
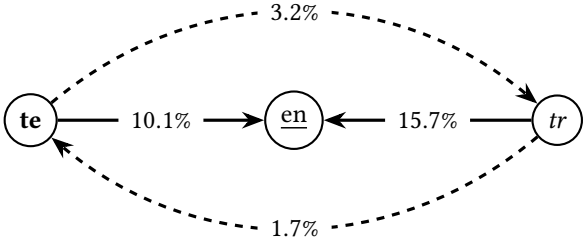

        \begin{tikzpicture} 
        \begin{scope}[every node/.style={circle,thick,draw}]
            \node (te) at (-1,0) {\textbf{te}};
            \node (en) at (2.5,0) {\underline{en}};
            \node (tr) at (6,0) {\emph{tr}};
        \end{scope}
        
        \begin{scope}[>={Stealth[black]},
                      every node/.style={fill=white,circle},
                      every edge/.style={draw=black,very thick}]
        
            \path [->] (te) edge node {$10.1\%$} (en);
            \path [->] (tr) edge node {$15.7\%$} (en);
            \path [->, dashed] (tr) edge[bend left=40] node {$1.7\%$} (te);
            \path [->, dashed] (te) edge[bend left=40] node {$3.2\%$} (tr); 
        
        \end{scope}
        \end{tikzpicture}
        \end{center}
        
        \setlength{\belowcaptionskip}{-10pt}
        
        \captionof{figure}{Overlap among the \textbf{target}, \emph{auxiliary} and \underline{pivot} language vocabularies.  Target $\rightleftharpoons$ auxiliary overlap numbers (dotted arrows) are measured before censoring the vocabularies to remove any overlap.  Directed edges indicate direction of asymmetric percent overlap $\frac{|V_{source} \cap V_{target}|}{|V_{source}|}$.}
        \label{fig:vocab_overlaps_2}

\end{minipage}\hfill
\begin{minipage}[b]{0.47\linewidth}

    \centering
    \begin{center}
    \begin{tabular}{|l|c|c|c|}
    \hline & \bf \emph{te+en} & \bf +\emph{tr} & \bf \% Change\\ \hline
    \textbf{Recall@1} & 17.0 & 17.6 & \textbf{+3.5\%} \\
    \hline
    \textbf{Recall@10} & 23.9 & 25.0 & \textbf{+4.6\%} \\
    \hline
    \textbf{Recall@20} & 26.3 & 27.7 & \textbf{+5.3\%} \\
    \hline
    \end{tabular}
    \end{center}
    \vspace{43pt} 
    \caption{Transfer improvement of Telugu (\emph{te}) in the presence of Turkish (\emph{tr}) data, even with no direct vocabulary overlap.  English (\emph{en}) serve as a pivot language.}
    \label{tab:xfer_transitive_2}
   \vspace{7pt} 

\end{minipage}
\end{table*}

        We design an experiment to test this hypothesis and determine whether transfer requires direct overlap between the vocabularies of two languages, or whether transfer can still happen even if the vocabularies of the two languages are disjoint.  For this experiment we focus on NSP and pick two languages with very little natural vocabulary overlap: Telugu (\emph{te}) and Turkish (\emph{tr}).  We train one model, \textbf{\emph{te+en}}, on sentence pairs drawn from the two languages Telugu (\emph{te}) and English (\emph{en}), in the same proportion as they occur naturally in the training data, and evaluate this model's performance on held out Telugu data.  We then similarly sample sentence pairs from an auxiliary language, Turkish (\emph{tr}), whose vocabulary has been censored to contain no overlap with the Telugu vocabulary.  Thus, there is no direct route by which the Turkish data can influence the embedding of the Telugu vocabulary (see Figure \ref{fig:vocab_overlaps_2} for details of all vocabulary overlaps). And yet, as shown in Table \ref{tab:xfer_transitive_2}, the addition of the Turkish data does improve the (\textbf{\emph{te+en+tr}}) model's performance on the Telugu data, despite there being no direct connection between the two languages. 

        While, by construction, there is no direct overlap between the target (\emph{te}) and auxiliary (\emph{tr}) language vocabularies, there is indirect overlap via the pivot language, English.  Thus, in a graphical sense, the influence of the auxiliary language is able to pass transitively through the chain of overlapping vocabularies of the pivot language to ultimately influence and improve the target language's performance.  If this link via the pivot language is removed, the performance of the target language is unchanged from its baseline, even in the presence of the auxiliary language.
        
    \begin{figure}[h]
        \centering

\begin{center}
\begin{tikzpicture}
\begin{scope}[every node/.style={circle,thick,draw}]
    \node (u1) at (-3, .75) {$u_1$};
    \node (u2) at (-1, .75) {$u_2$};
    \node (u3) at (1, .75) {$u_3$};
    \node (u4) at (3, .75) {$u_4$};

    \node (t1) at (-3, -.75) {$t_1$};
    \node (t2) at (-1, -.75) {$t_2$};
    \node (t3) at (1, -.75) {$t_3$};
    \node (t4) at (3, -.75) {$t_4$};
    
\end{scope}

\draw (-2,0) circle (1.85) (-2,1.85)  node [text=black,above] {$L_{auxiliary}$}
      (0,0) circle (1.85) (0,1.85)  node [text=black,above] {$L_{pivot}$}
      (2,0) circle (1.85) (2,1.85)  node [text=black,above] {$L_{target}$};

\begin{scope}[>={Stealth[black]},
              every node/.style={fill=white,circle},
              every edge/.style={draw=black,very thick}]

    \path [-] (u1) edge  (u2);
    \path [-] (u2) edge  (u3);
    \path [-] (u3) edge  (u4);
    
    \path [-] (u1) edge  (t1);
    
    \path [-] (t1) edge  (t2);
    \path [-] (t2) edge  (t3);
    \path [-] (t3) edge  (t4);
    
    \path [-, dashed] (u4) edge  (t4);

\end{scope}
\end{tikzpicture}

\caption{Language $L_{auxiliary}$ has tokens $u_1, u_2, t_1, t_2$ and evidence shows that $(u_1, t_1)$, $(u_1, u_2)$ and $(t_1, t_2)$ are synonyms (solid edges).  $L_{pivot}$ has tokens $u_2, u_3, t_2, t_3$ and evidence that $(u_2, u_3)$ and $(t_2, t_3)$ are synonyms.  Finally $L_{target}$ has tokens $u_3, u_4, t_3, t_4$ and evidence that $(u_3, u_4)$ and $(t_3, t_4)$ are synonyms.  Here evidence from $L_{auxiliary}$ influences $L_{target}$, via $L_{pivot}$, indicating that there is a relationship between $(u_4, t_4)$ (dashed edge), even though there is no direct lexical overlap among $L_{auxiliary}$ and $L_{target}$.}
\label{fig:indirect_xfer}

\end{center}
\end{figure}
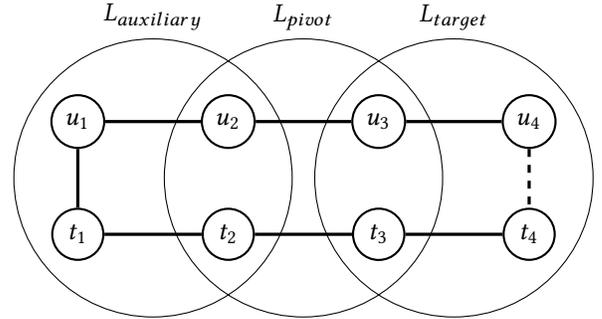

        Figure \ref{fig:indirect_xfer} shows a picture of how this indirect transfer might work, for a simplified setting where we must learn synonymy.  In this example, the auxiliary language is needed to discover the relationship between $(u_1, t_1)$, while the pivot language is needed to map $(u_2, u_3)$ and $(t_2, t_3)$.  Without both of these pieces of information, there would be no way to learn the implied edge between $(u_4, t_4)$ in the target language.  
        
        Generalizing from this example, we can see that the more auxiliary languages we have, the more potential information we have to learn about correlations among tokens and contexts.  Likewise, the more pivot languages we have, the more strongly connected our token graph is and the more paths we have for spreading the information learned from the auxiliary languages to the target language.  Of course, in the more realistic setting without artificial vocabulary censoring, languages can serve as both pivots and auxiliaries simultaneously, suggesting that the large scale multi-language transfer facilitated by our instance-based approach might be a critical component towards maximizing the amount of transfer enabled by these overlapping token networks.  In addition to these network effects, the added languages can also serve as regularizers, robustifying the models and leading to improved performance.
        
        This effect seems to be enabled and amplified by the large sample size and diverse vocabulary of the pivot language, even when the amount of auxiliary data is relatively small, and is most easily seen in code-mixing around proper nouns and names (see Table \ref{tab:kiev} for an example involving Russian (\emph{ru}), English (\emph{en}) and Ukrainian (\emph{uk})).

\begin{table*}[ht]
\renewcommand\thetable{4}
\begin{minipage}[b]{0.47\linewidth}
\centering

       
        \centering

    \begin{center}
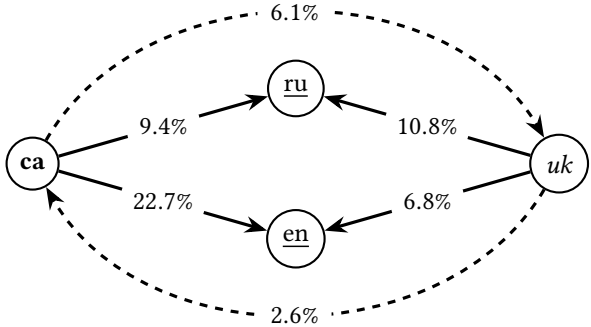

    \begin{tikzpicture} 
    \begin{scope}[every node/.style={circle,thick,draw}]
        \node (ca) at (-1,0) {\textbf{ca}};
        \node (ru) at (2.5,1) {\underline{ru}};
        \node (en) at (2.5,-1) {\underline{en}};
        \node (uk) at (6,0) {\emph{uk}};
    \end{scope}

    \begin{scope}[>={Stealth[black]},
              every node/.style={fill=white,circle},
              every edge/.style={draw=black,very thick}]

    \path [->] (ca) edge node {$9.4\%$} (ru);
    \path [->] (ca) edge node {$22.7\%$} (en);
    \path [->] (uk) edge node {$10.8\%$} (ru);
    \path [->] (uk) edge node {$6.8\%$} (en);
    \path [->, dashed] (uk) edge[bend left=60] node {$2.6\%$} (ca);
    \path [->, dashed] (ca) edge[bend left=60] node {$6.1\%$} (uk); 
    
    \end{scope}
    \end{tikzpicture}
    \end{center}
    

        \captionof{figure}{Overlap among the \textbf{target}, \emph{auxiliary} and \underline{pivot} language vocabularies.  Target $\rightleftharpoons$ auxiliary overlap numbers (dotted arrows) are measured before censoring the vocabularies to remove any overlap.  Directed edges indicate direction of asymmetric percent overlap $\frac{|V_{source} \cap V_{target}|}{|V_{source}|}$.}
        \label{fig:vocab_overlaps}

\end{minipage}\hfill
\begin{minipage}[b]{0.47\linewidth}

    \centering
    \begin{center}
        \begin{tabular}{|l|c|c|c|}
        \hline & \bf \emph{ca+en+ru} & \bf +\emph{uk} & \bf \% Change\\ \hline
        \textbf{Recall@1} & 10.6 & 11.3 & \textbf{+6.6\%} \\
        \hline
        \textbf{Recall@10} & 20.1 & 21.5 & \textbf{+7.0\%} \\
        \hline
        \textbf{Recall@20} & 24.3 & 25.6 & \textbf{+5.3\%} \\
        \hline
        \end{tabular}
        \end{center}
        \vspace{60pt} 
        \caption{Transfer improvement of Catalan (\emph{ca}) in the presence of Ukrainian (\emph{uk}) data, even with no direct vocabulary overlap.  English (\emph{en}) and Russian (\emph{ru}) serve as pivot languages.}
        \label{tab:xfer_transitive}
        \vspace{18pt}

\end{minipage}
\end{table*}

        To demonstrate that these results generalize across other language pairs, we show a similar effect with two other, similarly disjoint languages: Catalan (\emph{ca}) and Ukrainian (\emph{uk}).  This time we use two pivot languages: English (\emph{en}) and Russian (\emph{ru}).  We follow the same procedure as before, censoring the vocabularies of the four languages to ensure there is no overlap.  Figure \ref{fig:vocab_overlaps} shows the natural vocabulary overlaps between these four languages, while Table \ref{tab:xfer_transitive} summarizes the results of the experiment.  Once again we see that the addition of a seemingly unrelated language (Ukrainian) improves the performance on another language (Catalan), despite the fact that there is no direct vocabulary overlap among the languages.

       \begin{table}[h]
       \renewcommand\thetable{3}
        \begin{center}
        \begin{tabular}{|p{7.1cm}|}
        \hline

        \fontencoding{T2A}\selectfont

            \dots английского названия \underline{Киева}: в результате появилось слово \underline{Kyiv}\dots ими реже, чем \underline{Kiev}.\footnotemark \fontencoding{T1}\selectfont    \\
       \hline

        \end{tabular}
        \end{center}
        \vspace{10pt} 
        \caption{Example from Russian Wikipedia demonstrating code-mixing between Russian and English names for the city Kiev (underlined).  Translation: \emph{...English name of Kiev: as a result, the word Kyiv appeared...less often than Kiev}.}
        \label{tab:kiev}
        \end{table}

\section{Conclusion}

We have shown that cross-language instance-based transfer learning can significantly improve performance across all 35 languages tested on both the next sentence prediction and inverse cloze tasks, when formulated as multilingual deep retrieval problems.  We have identified sample size, task difficulty and vocabulary overlap as three factors that contribute to this technique's success, and demonstrated that transfer is possible even when there is only indirect transitive vocabulary overlap.  We have also shown that varying the mixture ratio between target and auxiliary data can further improve transfer performance. 

These very large-scale experiments with transfer between multiple languages have uncovered some
regularities not easily seen in smaller-scale experiments involving only two or three languages, or only one task.  This has been directly enabled by the highly scalable nature of the instance-based transfer method we describe, allowing more observations and the identification of subtler trends and relationships.

Looking forward, the results on direct vocabulary transfer show that transfer might be further improved by increasing the vocabulary overlap among languages, using features such as byte, character, subword and phoneme n-grams \cite{nguyen-chiang-2017-transfer, zhao-etal-2018-generalizing, wilson2008comparing}.  More surprisingly, the results on indirect vocabulary transfer suggest that researchers currently doing similar transfer experiments on a limited set of languages might see an immediate benefit from including a larger set of languages, even if they are seemingly unrelated, due to the transitive transfer effect.  These direct and transitive vocabulary overlap effects seem to be distinct from more indirect transfer effects identified in related work \cite{mueller-etal-2020-analysis,wu-dredze-2020-languages,karthikeyan2019cross}.  It is an open question if these kinds of transitive effects would also occur under the fine-tuning transfer approach.  Finally, further work is needed to investigate the impact that the improved transfer might have on downstream tasks.

\bibliographystyle{ACM-Reference-Format}
\bibliography{sample-base}

\appendix

\end{document}